\documentclass[10pt,twocolumn,letterpaper]{article}

\usepackage{wacv}
\usepackage{times}
\usepackage{epsfig}
\usepackage{graphicx}
\usepackage{amsmath}
\usepackage{amssymb}

\wacvfinalcopy 

\def\wacvPaperID{767} 

\ifwacvfinal\fi
\begin{document}

\title{Rotation-invariant Mixed Graphical Model Network \\for 2D Hand Pose Estimation}

\author{Deying Kong \\
University of California, Irvine\\
{\tt\small deyingk@uci.edu}
\and
Haoyu Ma \\
University of California, Irvine\\
{\tt\small haoyum3@uci.edu}
\and
Yifei Chen \\
Tencent Hippocrates Research Lab\\
{\tt\small dolphinchen@tencent.com}
\and
Xiaohui Xie \\
University of California, Irvine\\
{\tt\small xhx@uci.edu}
}

\maketitle
\ifwacvfinal\thispagestyle{empty}\fi

\begin{abstract}
In this paper, we propose a new architecture named Rotation-invariant Mixed Graphical Model Network (R-MGMN) to solve the problem of 2D hand pose estimation from a monocular RGB image. 
By integrating a rotation net, the R-MGMN is invariant to rotations of the hand in the image. It also has a pool of graphical models, from which a combination of graphical models could be selected, conditioning on the input image. Belief propagation is performed on each graphical model separately, generating a set of marginal distributions, which are taken as the confidence maps of hand keypoint positions. 
Final confidence maps are obtained by aggregating these confidence maps together. 
We evaluate the R-MGMN on two public hand pose datasets. Experiment results show our model outperforms the state-of-the-art algorithm which is widely used in 2D hand pose estimation by a noticeable margin. 
\end{abstract}


\section{Introduction}
Hands play a central role in almost all daily activities of human beings. Understanding hand pose is an essential task for many AI applications, such as gesture recognition, human-computer interaction~\cite{sridhar2015investigating}, and augmented/virtual reality~\cite{lee2009multithreaded,piumsomboon2013user}. 
The task of estimating hand pose has been investigated for decades, however, 
it still remains challenging due to the complicated articulation, high dexterity and severe self-occlusion. 

To address these problems, one possible way is to resort to multi-view camera systems~\cite{joo2015panoptic,panteleris2017back,simon2017hand}. However, such systems are expensive and not practical. 
Meanwhile, with the popularization of low-cost depth sensors in recent years, a large number of RGB-D based approaches have been proposed for 3D hand pose estimation~\cite{baek2018augmented,ge2016robust,ge2018point,mueller2017real,wan2019self,wan2018dense,yuan2018depth}. Nonetheless, RGB cameras are still the most popular and easily accessible devices. Researchers have started performing 3D hand pose estimation directly from RGB images~\cite{boukhayma20193d,cai2018weakly,ge20193d,iqbal2018hand,mueller2018ganerated,panteleris2018using,spurr2018cross,zimmermann2017learning}.
Many proposed approaches involve a two stage architecture, i.e., first performing 2D hand pose estimation and then lifting the estimated pose from 2D to 3D~\cite{boukhayma20193d,mueller2018ganerated,panteleris2018using,zimmermann2017learning}, which makes 2D hand pose estimation itself still an important task. In this paper, we investigate the problem of 2D hand pose estimation from a monocular RGB image. 

The research field of 2D hand pose estimation is related closely to that of human pose estimation. 
Spurred by developments in deep learning and large datasets publicly available~\cite{mueller2017real,simon2017hand}, deep convolutional neural network (DCNN)-based algorithms have make this field advance significantly. 
Convolutional Pose Machines (CPM)~\cite{wei2016convolutional} is one of the most popular and well known algorithms for human pose estimation, and it has been widely applied in 2D hand pose estimation~\cite{simon2017hand} yielding the state of the art performance.

Although the deep convolutional neural networks have the power to learn very good feature representations, they could only learn spatial relationships among joints or keypoints implicitly, which often results in joint inconsistency~\cite{ke2018multi,song2017thin}.  
To model the correlation among joints explicitly, several studies investigate the combination of Graphical Model (GM) and DCNN in pose estimation. In most of the studies~\cite{song2017thin,tompson2014joint,yang2016end}, a self-independent GM is imposed on top of the score maps regressed by DCNN. The parameters of the GM are learned during end-to-end training, then these parameters are fixed during prediction. However, pose can be varied in different scene, a fixed GM is unable to model diverse pose. This shortage could be even worse in hand pose estimation. In~\cite{chen2014articulated}, image-dependent pairwise potentials are introduced, however, the model does not support end-to-end training and the pairwise potential is restrained to quadratic function.

In this paper, we propose a novel architecture for 2D hand pose estimation from monocular RGB image, namely, the Rotation-invariant Mixed Graphical Model Network (R-MGMN). We argue that different hand shapes should be associated with different spatial relationships among hand keypoints, resulting to graphical models with different parameters. Also, a powerful graphical model should have the ability to capture the same shape of the hands when viewed from a different angle, i.e., the graphical model should be rotation-invariant. 

The proposed R-MGMN consists of four parts, i.e., a rotation net, a soft classifier and a pool which contains several different graphical models. 
The rotation net is inspired by the Spatial Transformer Networks~\cite{jaderberg2015spatial}.
The goal of the rotation net is to rotate the input image such that the hand would be in a canonical direction. Then, the soft classifier outputs a soft class assignment vector (which sums up to 1), representing the belief on possible shapes of the hand.
Meanwhile, the unary branch generates heatmaps which would be fed into the graphical models as unary functions.
After that, inference is performed via message passing on each graphical model separately. The inferred marginals are aggregated by weighted averaging, using the soft assignment vector. This procedure could be viewed as a soft selection of graphical models.
The final scoremap is obtained by rotating the aggregated marginal  backwards to align with the original coordinate of the input image.


We demonstrate the performance of the R-MGMN on two public datasets, the CMU Panoptic Dataset~\cite{simon2017hand} and the Large-scale 3D Multiview Hand Pose Dataset~\cite{gomez2017large}. Our approach outperforms the popularly used algorithm CPM by a noticeable margin on both datasets. Qualitative results indicate our model could alleviate geometric inconsistency among hand keypoints even when severe occlusion exists. 

The main contributions of this paper are summarized as follows:
\begin{itemize}
	\vspace{-0.2cm}
	\item We propose a new model named R-MGMN which combines graphical model and deep convolutional neural network efficiently. 
	\vspace{-0.2cm}
	\item Instead of only having one graphical model, the proposed  R-MGMN  has several independent graphical models which can be selected softly, depending on input image. And it could be trained end-to-end.

	\vspace{-0.2cm}
	\item Our R-MGMN could alleviate the spatial inconsistency among predicted hand keypoints greatly and outperform the popularly used CPM algorithm by a notable margin.
	
\end{itemize}

\section{Related Work}
\subsection{Human pose estimation from single RGB image}

Studies on hand pose estimation have been benefiting from that on human pose estimation for a long time. 
Since DeepPose~\cite{toshev2014deeppose} pioneered the application of DCNN in pose estimation, DCNN-based algorithms have dominated the field~\cite{gpa}. For example, the network proposed by Sun et al.~\cite{sun2019deep} has achieved the state-of-the-art score in many human pose estimation datasets~\cite{andriluka20142d, lin2014microsoft}. 
Early DCNN-based algorithms try to regress the 2D coordinates of keypoints~\cite{wang2016deeply, carreira2016human}. Later algorithms estimate keypoint heatmaps~\cite{wei2016convolutional,newell2016stacked,chu2017multi}, which usually achieve better performance. The main body of DCNN mainly adopts the high-to-low and low-to-high framework, optionally augmented with multi-scale fusion and intermediate supervision. However, the structure information among the body joints captured by DCNN is implicit. 
Some approaches try to learn extra information besides heatmaps of joint position to provide structural constraints, i.e. compound heatmaps~\cite{ke2018multi} and offset fields~\cite{papandreou2017towards}. Nonetheless, these methods still could not fully exploit structural information.

\subsection{Hand pose estimation}
Recently, most studies of hand pose focus on 3D hand pose estimation, which is much more challenging than body pose estimation, due to self-occlusion, dexterity and articulation of the hand. 
The mainstream approaches usually resort to either multi-view camera system~\cite{joo2015panoptic,panteleris2017back,simon2017hand} or depth data ~\cite{baek2018augmented,ge2018point,wan2019self,wan2018dense,yuan2018depth}.
Nevertheless, There is also a rich literature on 3D hand pose and reconstruction from single color image using deep neural networks~\cite{boukhayma20193d,cai2018weakly,ge20193d,iqbal2018hand,mueller2018ganerated,panteleris2018using,spurr2018cross,zimmermann2017learning}
. 
Some studies fit their 3D hand models from the estimated 2D joint locations~\cite{boukhayma20193d,mueller2018ganerated,panteleris2018using,zimmermann2017learning}. 
Thus the accuracy of 2D hand pose estimation has a great impact on the performance of 3D hand pose. 

Among a variety of DCNN-based models, CPM~\cite{wei2016convolutional} is commonly used in 2D hand pose estimation~\cite{simon2017hand,wang2018mask,zimmermann2017learning}. This architecture estimates the score maps via intermediate supervision and the most likely location is selected as the maximum confidence of the corresponding position in the confidence maps. In this paper, we choose CPM as the baseline for comparison with our proposed model. 

\subsection{Graphical model in pose estimation}
Graphical model has also been exploited in solving human pose estimation tasks. By using GM, spatial constraints among body parts can be modeled explicitly. 

Recently, there is also a trend to combine DCNN and GM for pose estimation~\cite{tompson2014joint,chen2014articulated,song2017thin,yang2016end}. The GM and the backbone DCNN are trained either independently or end-to-end via the combination of back propagation and message-passing. However, studies in this field usually apply a GM with fixed parameters, which limits its ability to model a variety of pose, especially in hand pose estimation. The most recent work in~\cite{kong2019adaptive} proposes to generate adaptive GM parameters conditioning on individual input images. 


\begin{figure*}[h]
\begin{center}
	\includegraphics[width=0.95\linewidth]{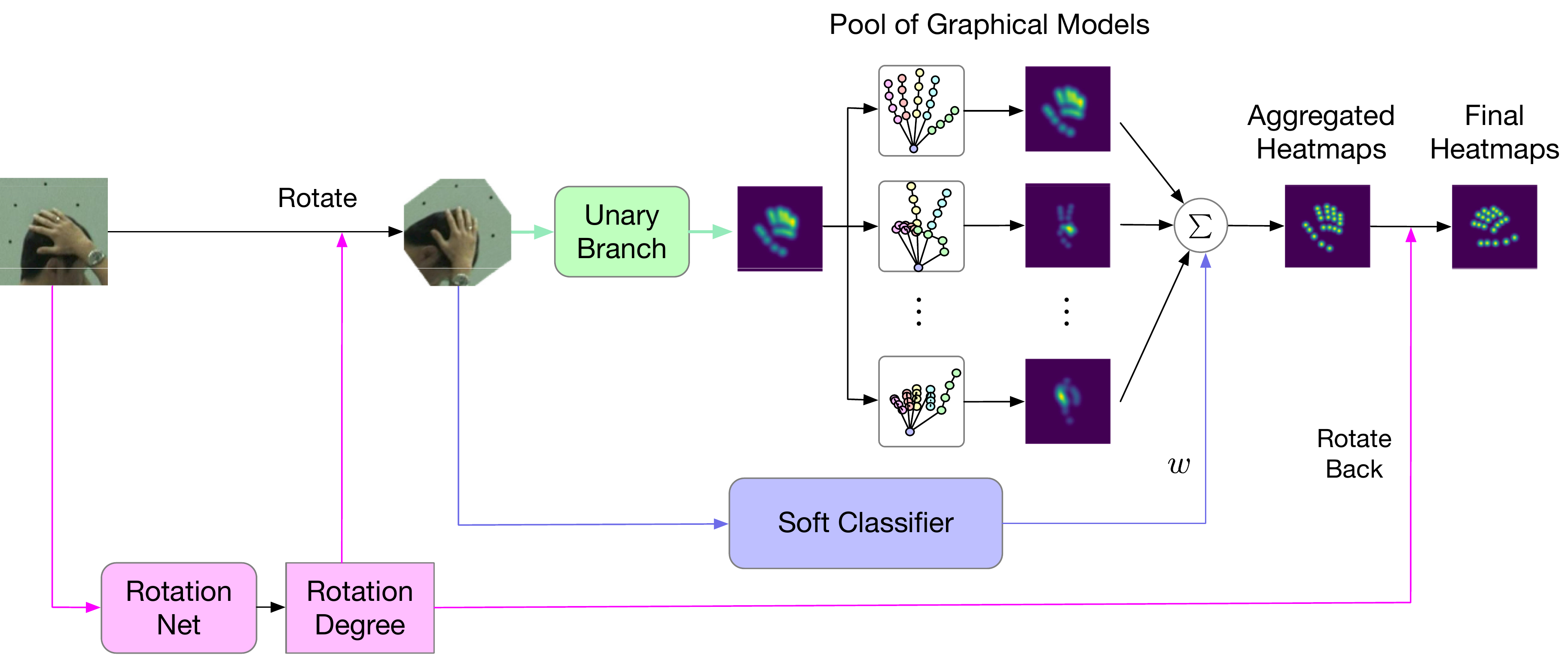}
\end{center}
   \caption{Pipeline overview of the proposed Rotation Mixture Graphical Model Network (R-MGMN).}
\label{fig:model_overview}
\end{figure*}

\section{Methodology}
\subsection{Basic pipeline}
The proposed Rotation-invariant Mixture Graphical Model Network (R-MGMN) mainly consists of four components, i.e., the rotation net, the soft classifier, the unary branch and the pool of graphical models, as shown in Fig.~\ref{fig:model_overview}. The pipeline of the MGMN is given as follows.

\begin{itemize}
	\vspace{-0.2cm}\item The rotation net regresses a rotation degree from the input image. 
	\vspace{-0.2cm}\item Then, using the obtained rotation degree, 
the image is rotated such that the hand in the image would be in a canonical direction (e.g. the hand is upright).
	\vspace{-0.2cm}\item After that two parallel branches follow.
	
	Branch 1:
    \begin{itemize}
    	\vspace{-0.2cm}\item A deep neural network referred to as unary branch is applied onto the rotated image. 
The output of the unary branch is a set of 2D heatmaps which represent the confidence of the hand keypoint positions.

        \vspace{-0.1cm}\item As unary potential functions, these 2D heatmaps are fed into the pool of graphical models. Each graphical model performs inference separately. Then, the pool outputs several sets of marginal probabilities of the keypoint positions. 
    \end{itemize}
    Branch 2:
    \begin{itemize}
    	\vspace{-0.2cm}\item The parallel branch contains a soft classifier which outputs a weight vector whose entries sum up to one.
    \end{itemize}

	\vspace{-0.2cm}\item Aggregated heatmaps are obtained as the weighted average of the marginal probabilities, using the weight vector.

   \vspace{-0.2cm} \item Rotate heatmaps backwards according to previous rotation degree.

\end{itemize}

\subsection{Model}

Our R-MGMN could be broken down into two parts:
\begin{itemize}
	\vspace{-0.15cm}\item The rotation part which controls the rotation of the image and the backward rotation of the heatmaps. 
	\vspace{-0.15cm}\item The MGMN, which performs handpose estimation on the rotated images.
\end{itemize}

\subsubsection{Image rotation}
The rotation angle $\alpha$ is regressed from the rotation net $\mathcal{RT}$ as 
\begin{equation}
\alpha = \mathcal{RT}(I; \theta_{rt}),
\label{eq:1.1}
\end{equation}
where $\theta_{rt}$ is the set of parameters of the rotation net, $I$ is the input image.
Then the rotated image is given by
\begin{equation}
I_{rt} = f_{rt}(I, \alpha),
\label{eq:1.2}
\end{equation}
where $f_{rt}$ is the rotation function.

\subsubsection{MGMN}

Given the rotated image $I_{rt}$, the handpose estimation problem could be formulated by using a graph and it could be solved via probabilistic tools. 


Let $\mathcal {V} = \{ v_1, v_2, \cdots, v_K\}$ denote the set of all the hand keypoints, each of which is associated with a random variable $x_i \in  \mathbb{R}^2$ representing its 2D position in image $I_{rt}$.
And let $\mathcal{E}$ represent the set of pairwise relationships among the keypoints in $\mathcal{V}$, to be more specific, $(i, j) \in \mathcal{E}$ if and only if ${v_i}$ and ${v_j}$ ($i < j$) are neighbours. Then we could define a graph $\mathcal{G} = (\mathcal{V}, \mathcal{E})$ with $\mathcal{V}$ being its vertices and $\mathcal{E}$ being the edges. A basic probabilistic model of the handpose task could be formulated by the following equation. 

\begin{equation}
p^{\texttt{basic}}(X | I_{rt}) = \prod_{v_i \in V} \phi(x_i|I_{rt}) \prod_{(j,k) \in \mathcal{E}}\psi(x_j, x_k|I_{rt}),
\label{eq:1}
\end{equation}
where $\phi(x_i) \in \mathbb{R}$ is usually called the unary function, $\psi(x_j, x_k) \in  \mathbb{R}$  is the pairwise function and $X$ denotes the positions of all hand keypoints, i.e., $ X =  (x_1, x_2, \cdots, x_K)$ .

The naive model in Eq. (\ref{eq:1}) could be generalized to a mixed graphical model as 
\begin{equation}
p(X| I_{rt}) = \sum_{l=1}^{L} w _l \prod_{v_i \in V} \phi_l(x_i | I_{rt}) \prod_{(j,k) \in \mathcal{E}}\psi_l(x_j, x_k | I_{rt}),
\label{eq:2}
\end{equation}
where $L$ graphical models are aggregated together, $w_l$ is the weight corresponding to the $l$-th graphical model.

Our proposed MGMN is obtained when the same unary function is shared for all $L$ graphical models, i.e.,
\begin{equation}
\phi_l(x_i|I) =\eta(x_i|I) , \; l = 1,2,\cdots, L
\label{eq:3}
\end{equation}
where $\eta(x_i|I) $ is the output of the unary branch $\mathcal {U} $ with parameters $\theta_{ {u} }$ in Fig.~\ref{fig:model_overview}, 

\begin{equation}
\eta(x_i|I)=\mathcal{U}(I, \theta_{ {u} }).
\label{eq:3}
\end{equation}

The marginal probability $p(x_i \vert I_{rt})$ could be calculated by summing up the marginal probabilities $p_l(x_i \vert I_{rt})$ of each individual graphical models, as validated by the following equation,
\begin{align}
&p(x_i \vert I_{rt})
= \sum_{\sim x_i} p(X \vert I_{rt}) \\
=&\sum_{\sim x_i}  \sum_{l=1}^{L} w _l \prod_{v_i \in V} \phi_l(x_i | I_{rt}) \prod_{(j,k) \in \mathcal{E}}\psi_l(x_j, x_k | I_{rt})\\
=&\sum_{l=1}^{L} w _l \sum_{\sim x_i}  \prod_{v_i \in V} \phi_l(x_i | I_{rt}) \prod_{(j,k) \in \mathcal{E}}\psi_l(x_j, x_k | I_{rt})\\
=&\sum_{l=1}^{L} w _l \; p_l(x_i \vert I_{rt}),
\label{eq:6}
\end{align}
where $ \sum\limits_{\sim x_i} $ means to summing over all $x_k, k=1,2,\cdots, K$ except $x_i$.
The marginal $p_l(x_i \vert I_{rt})$ of each graphical model could be calculated exactly or approximately using message passing efficiently.

The marginal $p(x_i \vert I_{rt})$ could be taken as a confidence of the joint $v_i$ being located at the specific position. For each keypoint $v_i$, a confidence map or score map $S_i$, which is a 2D matrix,  could be constructed by assigning the $(m, n)$-th entry of $S_i$ to be
\begin{equation}
S_i[m,n] = p(x_i = (m,n) \vert I_{rt}) ,
\end{equation}
where $(m,n)$ is the 2D coordinate.

\subsubsection{Inverse Rotation of Confidence Maps}

The final confidence map of the keypoint $v_i$'s position is given by
\begin{equation}
S^{\texttt F}_i = f_{rt}(S_i, - \alpha),
\end{equation}
where $\alpha$ is given by the rotation net as in Eq.~(\ref{eq:1.1}).

The predicted position of keypoint ${v_i}$ is obtained by maximizing the confidence map, as 

\begin{equation}
x_i^{*} = (m^*, n^*) = \operatorname*{argmax}_{m,n}\; S^{\texttt F}_i [m, n].
\label{eq:1.2}
\end{equation}

\subsection{Detailed structure of the R-MGMN}
In this subsection, we would describe the detailed structure of each component of the R-MGMN.

\subsubsection{Rotation Net}

\begin{figure}[h]
\begin{center}
   \includegraphics[width=0.8\linewidth]{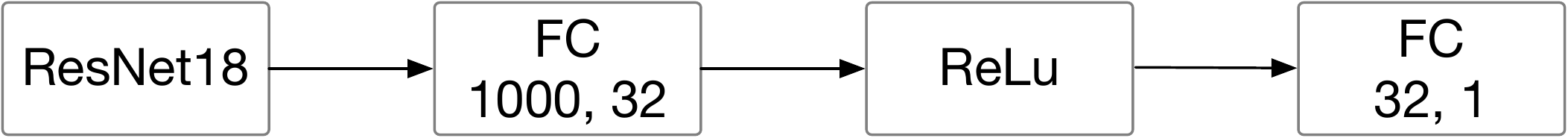}
\end{center}
   \caption{Configuration of the rotation net.}
\label{fig:rotation_net}
\end{figure}

The rotation net consists of a ResNet18 and two additional layers to regress the rotation degree $\alpha$, as shown in Fig.~\ref{fig:rotation_net}.

The output of ResNet18 is a 1000-dimensional vector, then it's fed into two fully connected (FC) layers with a ReLu function in between. Finally, a scalar representing the rotation degree is obtained. 

\subsubsection{Unary Branch}

The Convolutional Pose Machine (CPM) is adopted as our unary branch. 
To be more specific, we follow the same architecture used in~\cite{simon2017hand}.
The convolutional stages of a pre-initialized
VGG-19 network up to conv4\_4 are utilized as a feature extractor. Then, six cascaded stages are deployed to regress the confidence maps repeatedly.  Moreover, as in ~\cite{cao2018openpose}, convolutions of kernel size 7 are replaced with 3 layers of convolutions of kernel 3 which are concatenated at their end. 

\subsubsection{Soft Classifier}
\begin{figure}[h]
\begin{center}
   \includegraphics[width=0.4\linewidth]{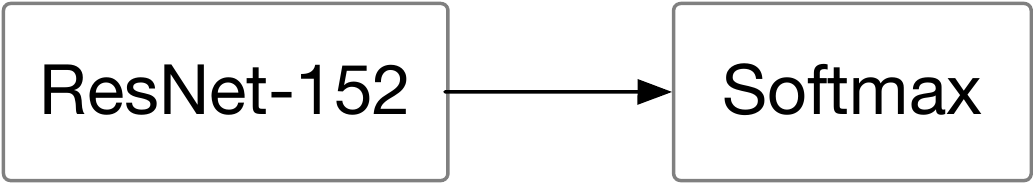}
\end{center}
   \caption{Configuration of the soft classifier.}
\label{fig:soft_classifier}
\end{figure}

For the soft classifier we adopt the ResNet-152 followed by a softmax layer as in Fig.~\ref{fig:soft_classifier}. 
The output dimension of the ResNet-152 is set to be 20, which means we would like to expect there are 
20 clusters among the hands.

\subsubsection{Pool of Graphical Models}
There are $L=20$ tree-structured graphical models integrated in the pool of graphical models. Each of the graphical model shares the same structure, but every single graphical model is associated with a different set of parameters. Marginal probabilities are inferred on each individual graphical model, and then aggregated via a weight vector which comes from the soft classifier.

{\bf Belief propagation.}
Sum-product message passing is a well known algorithm for performing inference on graphical models.
It could calculate marginals of the random variables efficiently. 
During the inference, vertices on the graph receive messages from and send messages to their neighbors 
iteratively, as in the following equation,
\begin{equation}
m_{ij} (x_j) = \sum_{x_i} \; \varphi_{i,j}(x_i, x_j) \phi_i(x_i) \prod_{k \in \texttt{Nbd}(i)\backslash j} m_{ki}(x_i)\, ,
\label{eq:14}
\end{equation}
where $m_{ij} \in \mathbb{R}$ is the message sent from vertex $v_i$ to vertex $v_j$, which is the belief from the vertex $v_i$ on the position of the 
$j$-th keypoint.

The message passing process in the above equation would be performed several iterations until convergence or satisfaction of some other stop criteria. The estimated marginal distribution $\hat{p_i} (x_i)$ is given by 
\begin{align}
 \hat{p_i} (x_i) &\propto \phi_i(x_i) \prod_{k \in \texttt{Nbd}(i)} m_{ki}(x_i) \\
                 & = \frac{1}{Z'} \phi_i(x_i) \prod_{k \in \texttt{Nbd}(i)} m_{ki}(x_i),
\label{eq:6}
\end{align}
where $Z'$ is a normalization term such that the probabilities sum up to 1. 

\begin{figure}[h]
\begin{center}
   \includegraphics[width=0.4\linewidth]{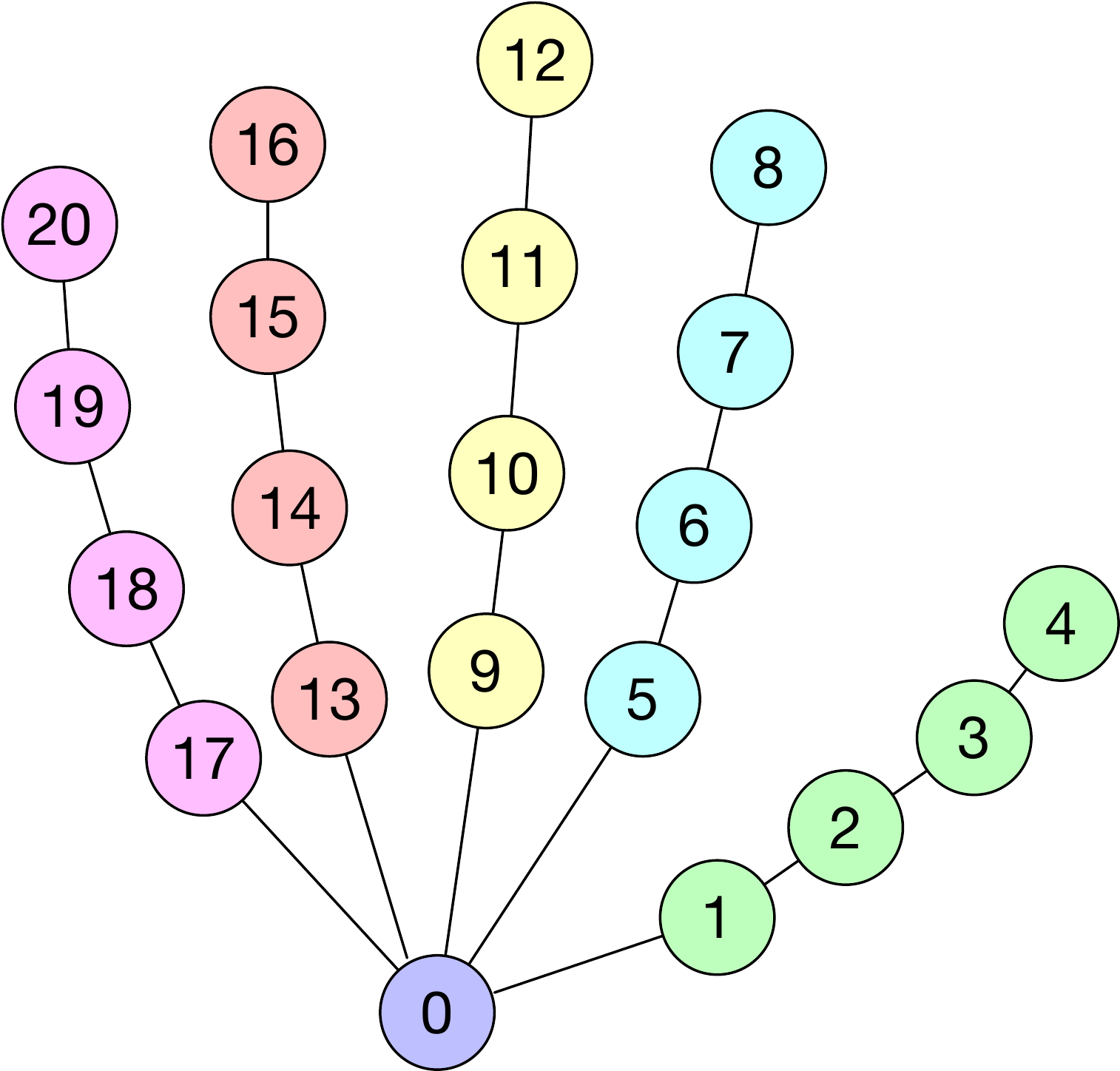}
\end{center}
   \caption{Tree-structured graphical model for hand keypoints.}
\label{fig:hand_model}
\end{figure}

{\bf Message passing on tree-structured graphs.}
When the graph is tree-structured, the estimated marginal equals the exact marginal. In our R-MGMN, tree-structured models are utilized, as illustrated in Fig.~\ref{fig:hand_model}. Each branch consisting of four same-colored circles corresponds to a single finger.

By using a tree-structure, exact marginals could be inferred very efficiently by only two passes of message passing. In the first step, starting from the leaf nodes, variables pass messages sequentially towards the root node. Then in the second step, messages are passed sequentially towards the leaf nodes, beginning at the root node. 

{\bf Message passing as 2D convolution.} For each iteration, the message $m_{ij}(x_j)$ in Eq.~(\ref{eq:14}) could be rewritten as
\begin{equation}
m_{ij} (x_j) = \sum_{x_i} \; \varphi_{i,j}(x_i, x_j) h_i(x_i), \\
\label{eq:17}
\end{equation}
where
\begin{equation}
h_i(x_i) \triangleq \phi _i(x_i) \prod_{k \in{\texttt Nbd(i)}\backslash j} m_{ki}(x_i).
\label{eq:18}
\end{equation}
If the pairwise potential function $\varphi_{i,j}(x_i, x_j)$ only depends on the relative position between the two neighboring keypoints, i.e.,
\begin{equation}
\varphi_{i,j}(x_i, x_j) = \gamma_{i,j}(x_i - x_j) \,.
\label{eq:19}
\end{equation}
By compacting $m_{i,j}(\cdot)$ and $h_i(\cdot)$ into 2D matrices $M_{ij}$ and $H_i$  (this is reasonable since $x_i$ corresponds to a 2D location), Eq.~(\ref{eq:17}) is transformed to
\begin{equation}
M_{ij}= \Gamma ^{i,j} \circledast H_i \,,
\label{eq:20}
\end{equation}
where $\Gamma ^{i,j}$ is a 2D matrix encoding the pairwise potential function $\gamma_{i,j}(x_i - x_j)$, and the notation $\circledast$ denotes convolution.

Thus, the set of parameters for each graphical model is given by
\begin{equation}
\Theta^{\texttt{gm}} = \{\Gamma ^{i,j} \vert (i,j) \in \mathcal{E} \; \mbox{or}\;  (j,i) \in \mathcal{E} \}  \,.
\label{eq:21}
\end{equation}
The whole set of parameters of the pool of graphical models are
\begin{equation}
\Theta^{\texttt{GM}} = \{ \Theta^{\texttt{gm}}_{l}\; \vert \; l=1,2,\cdots, L \}\, ,
\label{eq:22}
\end{equation}
where $\Theta^{\texttt{gm}}_{l_1}$ is independent of  $\Theta^{\texttt{gm}}_{l_2}$ for $l_1 \neq l_2$.

\section{Learning}
Since our R-MGMN contains several components, we follow a step-by-step training procedure. First, the rotation net is trained. Then, while keeping the rotation net fixed, we train the unary branch and the soft classifier separately. After that, the parameters of graphical models are learned while keeping other parts frozen. Finally, the whole R-MGMN is jointly trained. More details are given as following.

\subsection{Train Rotation Net}

\begin{figure}[h]
\begin{minipage}{0.43\textwidth}
\includegraphics[width=.2\textwidth]{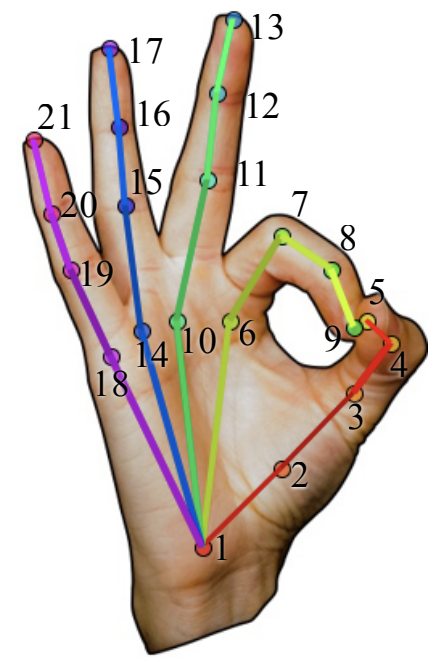}\hfill%
\includegraphics[width=.3\textwidth]{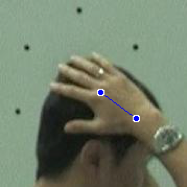}\hfill%
\includegraphics[width=.3\textwidth]{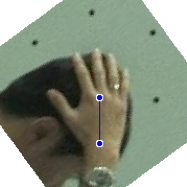}\hfill%
\end{minipage}

\vspace{1em}

\caption{Illustration of the rotation. Left image courtesy to~\cite{simon2017hand} .}
\label{fig:hand_rotation}
\end{figure}

The rotation net is trained alone in the first phase of the training. The aim of the rotation net is to rotate the input image such that the hand in the resulted image is upwards, i.e., the directional line connecting the 1-st keypoint and the 10-th keypoint is pointing upwards as illustrated in Fig.~\ref{fig:hand_rotation}.

Almost no public dataset provides the ground truth rotation degree directly, however, it could be obtained easily given the ground truth positions of hand keypoints. The ground truth rotation degree $\alpha ^{\texttt{*}}$ could be derived by calculating the directional angle between the vector $v_1$ and $v_2$, where 
\begin{equation}
    v_1 = x_{10} - x_1,
\label{eq:add 1}
\end{equation}
with ${x_{10}}$ and ${x_1}$ representing the positions of the keypoints, and $v_2$ is the unit vector whose direction is vertically upwards.

During training, squared error is used for the loss function, which is
\begin{equation}
L^{\texttt{rn}} = (\alpha - \alpha ^*)^2\, ,
\label{eq:23}
\end{equation}
where $\alpha$ is the regressed rotation degree from the rotation net.

\subsection{Train Unary Branch}
The unary branch is trained with the help of the rotation net, while the rotation net is fixed during this training phase. 
The unary branch is actually the convolutional pose machine, which produces and refines the confidence maps repeatedly. 
Rotated image is fed into the unary branch, which outputs a set of confidence maps. These confidence maps are then rotated back so as to be aligned with the original coordinate of the input image before the rotation net.

Denote $S^t_k\in \mathbb{R}^{h_u \times w_u}$ as the aligned output confidence map of the $k$-th keypoint at the $t$-th stage of the unary branch, the loss function used in this training phase is designed as 
\begin{equation}
L^{\texttt{unary}}=\sum_{t=1}^{T}  \sum_{k=1}^{21}||S_k^{t} - S_k^* ||_F^2  \, ,
\label{eq:24}
\end{equation}
where $T$ is the number of stages in the unary branch, $S_k^* \in \mathbb{R}^{h_u \times w_u}$ is the ground truth confidence map of the $k$-th keypoint, and $||\cdot||_F$ represents the Frobenius norm. The ground truth $S_k^*$ is obtained by by putting a Gaussian peak at the keypoint's ground truth location.

\subsection{Train Soft Classifier}
Again, there is no ground truth class label for the classification subtask. Thus, we resort to unsupervised learning, especially the K-means clustering algorithm. To be fair, only training dataset is utilized in this phase.

Given the pretrained rotation net in the first phase, we rotate the images and the keypoints' position labels according to the estimated rotation degrees. Then, the K-means algorithm is applied on the rotated images. The feature vector used in K-means is obtained by concatenating the relative positions of neighbouring keypoints. The number of the clusters is set to be 20.

The training set is further split into 70/30, on which the soft classifier is trained on. Standard cross entropy is used for the loss function.

\subsection{Train Graphical Model Parameters}
Keeping all the other parts fixed, in this phase, we only train the parameters of the graphical models, with the whole R-MGMN. The loss function is 
\begin{equation}
L^{\texttt{GM}} =  \sum_{k=1}^{21}||\tilde S_k - \tilde S_k^* ||_F^2,
\end{equation}
where $\tilde S_k \in \mathbb{R}^{ h_o \times w_o}$ is the $k$-th channel of the  output of the R-MGMN. Since the confidence map $\tilde S_k$ is actually a normalized probability distribution, the ground truth $\tilde S_k^* \in \mathbb{R}^{ h_o \times w_o}$ is also normalized ($\tilde S_k^*$ is the normalized version of $ S_k^*$ from Eq.~(\ref{eq:24})).

\subsection{Jointly Train All the Parameters}
For the last phase, we use the same loss function as that in training the graphical model paramters,

\begin{equation}
L^{\texttt{Joint}} =  L^{\texttt{GM}}\,.
\end{equation}

\begin{figure*}[h]

\includegraphics[width=.48\textwidth]{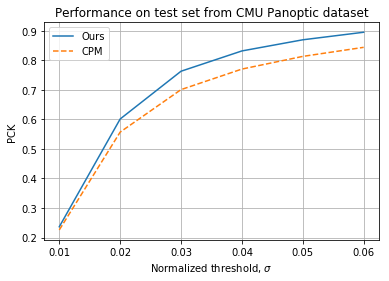}\hfill%
\includegraphics[width=.48\textwidth]{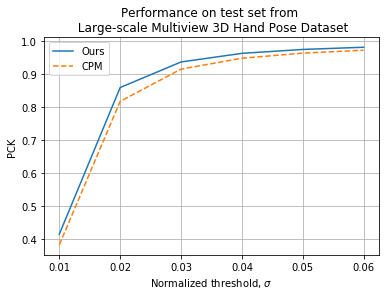}\hfill%

  \caption{PCK performance on two public datasets.}
  \label{fig:pcks}
\end{figure*}

\begin{table*}[t!]
\begin{center}
\begin{tabular}{ c | c c c c c c|c }
\hline
 Threshold of PCK, $\sigma$& 0.01& 0.02& 0.03& 0.04& 0.05& 0.06& mPCK\\
 \hline
  \multicolumn{8}{c}{CMU Panoptic Hand Dataset}\\
  \hline
 CPM Baseline (\%) & 22.60& 55.69& 70.06& 77.01& 81.30& 84.36& 65.17\\ 
 \hline
 Ours           & 23.67& 60.12& 76.28& 83.14& 86.91& 89.47& 69.93 \\  
 \hline
 Improvement          & 1.07& 4.43& \textbf{6.22}& \textbf{6.13}& \textbf{5.61}& \textbf{5.11}& \textbf{4.76}\\
 \hline
\hline

 \multicolumn{8}{c}{Large-scale Multiview 3D Hand Pose Dataset}\\
 \hline
  CPM Baseline (\%) & 38.27& 81.78& 91.54& 94.84& 96.39& 97.27& 83.35\\
 \hline
  Ours           & 41.51& 85.97&93.71& 96.33& 97.51& 98.17 & 85.53\\ 
  \hline
 Improvement     & \textbf{3.24}& \textbf{4.19}& 2.17& 1.49& 1.12& 0.90& 2.18\\
 \hline
\end{tabular}
\caption{Detailed numerical results of PCK performance.}
\label{table:pck}
\end{center}
\end{table*}

\section{Experiments}

We verify our approach on two public handpose datasets, i.e., the CMU Panoptic Hand Dataset (CMU Panoptic)~\cite{simon2017hand} and the Large-scale Multiview 3D Hand Pose Dataset (Large-scale 3D)~\cite{Francisco2017}. A comprehensive analysis of the proposed model is also carried out.
\subsection{Experimental settings}
\subsubsection{Datasets.} The CMU Panoptic dataset contains 14817 annotations of hand images while the Large-scale 3D dataset contains 82760 anotations in total. 
The Large-scale 3D dataset provides a simple interface to generate 2D labels from the 3D labels which come with the dataset.
For both datasets, we split them into training set (70\%), validation set (15\%) and test set (15\%). Since we focus on handpose estimation in this paper, we crop image patches of annotated hands off the original images, thus leaving out the task of hand detection. A square bounding box which is 2.2 times the size of the hand is used during the cropping.

\subsubsection{Evaluation metric.}
Probability of Correct Keypoint (PCK)~\cite{simon2017hand} is a popular metric, which is defined as the probability that a predicted keypoint is within a distance threshold $\sigma$ of its true location. In this paper, we use normalized threshold $\sigma$ with respect to the size of hand bounding box, and mean PCK (mPCK) with $\sigma = \{0.01, 0.02,0.03,0.04,0.05,0.06\}$.

\subsubsection{Implementation Details.}
All input images are resized to $368 \times 368 $, then scaled to [0,1] and further normalized using mean of (0.485, 0.456, 0.406) and standard derivation of (0.229, 0.224, 0.225).
Batch size is set to 32 for all training phases. Adam is used as the optimizer, and the initial learning rate is set to be $lr=$1e-4 for each training phase.
The rotation net is only trained for 6 epochs at the fist training phase. 
With best models being selected basing on the performance of the validation set, the unary branch and soft classifier are both trained for 100 epochs, after which the parameters of graphical models are trained for 40 epochs, and finally the whole network are trained end-to-end for 150 epochs.

\subsection{Results}
The PCK performance of our proposed model on two public datasets, i.e., the CMU Panoptic dataset and the Large-scale 3D dataset,  are shown in Fig.~\ref{fig:pcks}. It is seen that our model outperforms the CPM consistently on both datasets. Detailed numerical results are given in Table~\ref{table:pck}.

\begin{table*}[]
\begin{center}
\begin{tabular}{ c | c c c c c c|c||c }
\hline
 Threshold of PCK, $\sigma$& 0.01& 0.02& 0.03& 0.04& 0.05& 0.06& mPCK & improvement\\
 \hline
  \hline
 CPM Baseline (\%) & 22.60& 55.69& 70.06& 77.01& 81.30& 84.36& 65.17 & -\\ 
 \hline
  CPM + Single GM           & 22.58& 55.78& 70.14& 77.05& 81.34& 84.41& 65.21 & 0.04\\  
 \hline
  CPM + Mixture of GMs          & 23.39& 57.53& 71.95& 78.49& 82.28& 85.02& 66.44 & 1.27   \\  
 \hline
  Rotaion + CPM $^1$          & 22.70& 57.91& 72.95& 79.94& 83.90& 86.71& 67.35 & 2.18\\   
 \hline
  Rotaion + CPM $^2$          & 21.97& 57.59& 74.53& 81.98& 86.21& 88.83& 68.52 & 3.35\\   
 \hline
 R-MGMN           & 23.67& 60.12& 76.28& 83.14& 86.91& 89.47& 69.93 & 4.76 \\  
 \hline
\hline
\end{tabular}
\caption{Numerical results for ablation study on CMU Panoptic Hand Dataset.}
\label{table:ablation_study}
\end{center}
\end{table*}

\begin{figure*}[h]
\begin{center}
\includegraphics[width=0.9\linewidth]{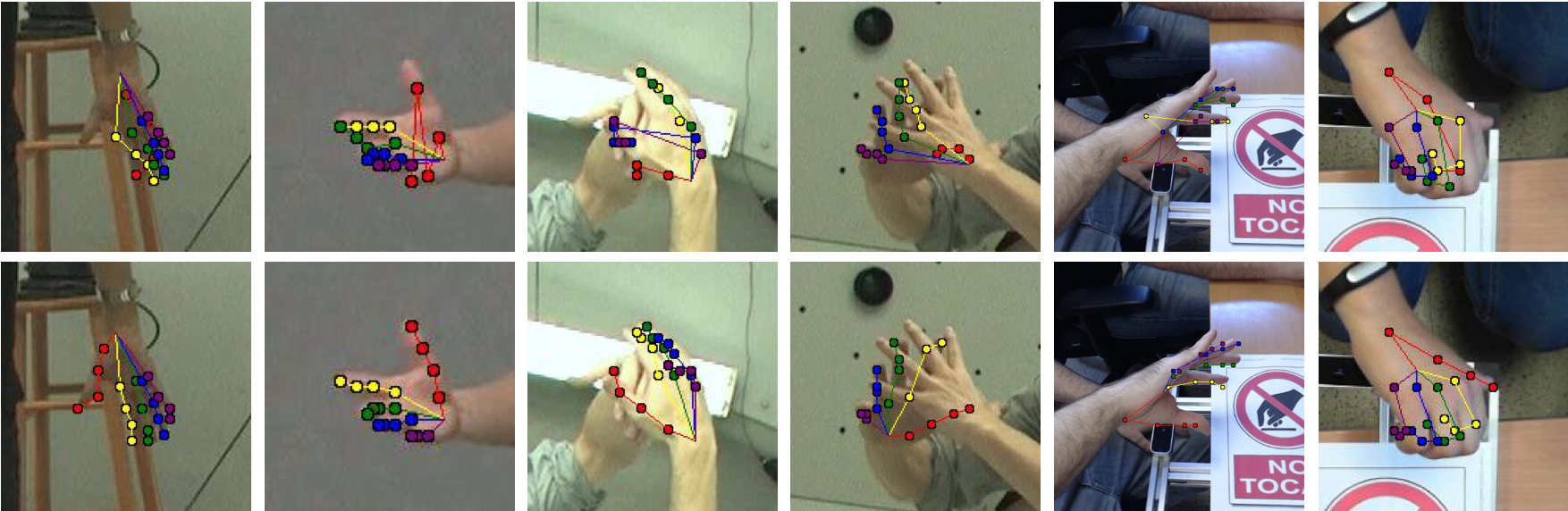}
\end{center}
\caption{Qualitative results. First row: CPM. Second row: our model.}
\label{fig:qualitative}
\end{figure*}

On CMU Panoptic dataset, our model achieves a significant PCK improvement comparing to CPM. An absolute improvement of 6.22 percent is observed at threshold $\sigma = 0.03$. In average, the mPCK is improved by 4.76 percent. The experiment result on Large-scale 3D dataset also validates the advantage of our model. At threshold of $\sigma = 0.02$, there is a 4.19 percent improvement in PCK.

The reason why the improvement on Large-scale 3D dataset is not as much as that on the CMU Panoptic dataset, probably lies in the fact that annotation settings of these two datasets are slightly different. 
In Large-scale 3D dataset, the center of the palm is considered as the root keypoint instead of the wrist. This would cause the reference vector $v_1$ in Eq.~(\ref{eq:add 1}) to be relatively short, which in turn would cause the calculated rotation degree to be prone to erroneous when noise exists.


Qualitive results are shown in Fig.~\ref{fig:qualitative}. Images in the top row are the predicted results by CPM, while the bottom row corresponds to the prediction of our model. The results show that our proposed R-MGNM could greatly reinforce the keypoints consistency, and generate much more reasonable predictions than CPM.

Our model succeed to predict well even if the hand is severely occluded, as in the 4-th column in Fig.~\ref{fig:qualitative}.  
In this example, half of the right hand is occluded by the left hand. The CPM fails to recover many of the keypoints. However, our R-MGMN correctly recovers the index finger and thumb, even they are totally occluded.

\subsection{Ablation study}
To understand the proposed model, ablation study is also performed. Several experiments are conducted as follows.
\begin{itemize}
	\vspace{-0.18cm}\item CPM+Single GM. In this experiment, we only keep the unary branch and one single graphical model from the R-MGMN. Both the rotation net and the soft classifier are removed. 
	\vspace{-0.18cm}\item CPM+Mixture of GMs. The rotation net is removed from the R-MGMN.
	\vspace{-0.18cm}\item Rotaion+CPM$^1$. Only keep the rotation net and the unary branch, jointly trained using the loss function in Eq.~(\ref{eq:24}).
	\vspace{-0.18cm}\item Rotaion+CPM$^2$. First train the rotation net, then jointly the train the rotation net and the unary branch.
\end{itemize}
All of the above experiments support end-to-end training.  Numerical results are given in Table~\ref{table:ablation_study}. As indicated by the results, adding a single graphical model on top of CPM has very little effect on the PCK performance. By adding a mixture of graphical models, there is an improvement of 1.28 percent in mPCK. Properly tuned, the rotation net would help improve the performance by 3.35 percent.
By integrating the rotation net and the mixture of graphical models together into our R-MGMN, final improvement of 4.76 percent is achieved.

\section{Conclusion}
A new architecture called Rotation-invariant Mixed Graphical Model Network (R-MGMN) is proposed in this paper. The R-MGMN combines the graphical model and deep convolutional neural network in a new way, where a pool of graphical models could be selected softly depending on input image. The R-MGMN could be trained end-to-end. Experiment results validate that the proposed R-MGMN outperforms the widely used CPM algorithm on two public datasets. Ablation study is also performed to see the functionality of each part of the R-MGMN model.


\clearpage
{\small
\bibliographystyle{ieee}
\bibliography{egbib}

\begin{thebibliography}{10}\itemsep=-1pt

\bibitem{andriluka20142d}
M.~Andriluka, L.~Pishchulin, P.~Gehler, and B.~Schiele.
\newblock 2d human pose estimation: New benchmark and state of the art
  analysis.
\newblock In {\em Proceedings of the IEEE Conference on computer Vision and
  Pattern Recognition}, pages 3686--3693, 2014.

\bibitem{baek2018augmented}
S.~Baek, K.~In~Kim, and T.-K. Kim.
\newblock Augmented skeleton space transfer for depth-based hand pose
  estimation.
\newblock In {\em Proceedings of the IEEE Conference on Computer Vision and
  Pattern Recognition}, pages 8330--8339, 2018.

\bibitem{boukhayma20193d}
A.~Boukhayma, R.~d. Bem, and P.~H. Torr.
\newblock 3d hand shape and pose from images in the wild.
\newblock In {\em Proceedings of the IEEE Conference on Computer Vision and
  Pattern Recognition}, pages 10843--10852, 2019.

\bibitem{cai2018weakly}
Y.~Cai, L.~Ge, J.~Cai, and J.~Yuan.
\newblock Weakly-supervised 3d hand pose estimation from monocular rgb images.
\newblock In {\em Proceedings of the European Conference on Computer Vision
  (ECCV)}, pages 666--682, 2018.

\bibitem{cao2018openpose}
Z.~Cao, G.~Hidalgo, T.~Simon, S.-E. Wei, and Y.~Sheikh.
\newblock Openpose: realtime multi-person 2d pose estimation using part
  affinity fields.
\newblock {\em arXiv preprint arXiv:1812.08008}, 2018.

\bibitem{carreira2016human}
J.~Carreira, P.~Agrawal, K.~Fragkiadaki, and J.~Malik.
\newblock Human pose estimation with iterative error feedback.
\newblock In {\em Proceedings of the IEEE conference on computer vision and
  pattern recognition}, pages 4733--4742, 2016.

\bibitem{chen2014articulated}
X.~Chen and A.~L. Yuille.
\newblock Articulated pose estimation by a graphical model with image dependent
  pairwise relations.
\newblock In {\em Advances in neural information processing systems}, pages
  1736--1744, 2014.

\bibitem{chu2017multi}
X.~Chu, W.~Yang, W.~Ouyang, C.~Ma, A.~L. Yuille, and X.~Wang.
\newblock Multi-context attention for human pose estimation.
\newblock In {\em Proceedings of the IEEE Conference on Computer Vision and
  Pattern Recognition}, pages 1831--1840, 2017.

\bibitem{Francisco2017}
S.~O.-E. Francisco Gomez-Donoso and M.~Cazorla.
\newblock Large-scale multiview 3d hand pose dataset.
\newblock {\em ArXiv e-prints 1707.03742}, 2017.

\bibitem{ge2016robust}
L.~Ge, H.~Liang, J.~Yuan, and D.~Thalmann.
\newblock Robust 3d hand pose estimation in single depth images: from
  single-view cnn to multi-view cnns.
\newblock In {\em Proceedings of the IEEE conference on computer vision and
  pattern recognition}, pages 3593--3601, 2016.

\bibitem{ge20193d}
L.~Ge, Z.~Ren, Y.~Li, Z.~Xue, Y.~Wang, J.~Cai, and J.~Yuan.
\newblock 3d hand shape and pose estimation from a single rgb image.
\newblock In {\em Proceedings of the IEEE Conference on Computer Vision and
  Pattern Recognition}, pages 10833--10842, 2019.

\bibitem{ge2018point}
L.~Ge, Z.~Ren, and J.~Yuan.
\newblock Point-to-point regression pointnet for 3d hand pose estimation.
\newblock In {\em Proceedings of the European Conference on Computer Vision
  (ECCV)}, pages 475--491, 2018.

\bibitem{gomez2017large}
F.~Gomez-Donoso, S.~Orts-Escolano, and M.~Cazorla.
\newblock Large-scale multiview 3d hand pose dataset.
\newblock {\em arXiv preprint arXiv:1707.03742}, 2017.

\bibitem{iqbal2018hand}
U.~Iqbal, P.~Molchanov, T.~Breuel Juergen~Gall, and J.~Kautz.
\newblock Hand pose estimation via latent 2.5 d heatmap regression.
\newblock In {\em Proceedings of the European Conference on Computer Vision
  (ECCV)}, pages 118--134, 2018.

\bibitem{jaderberg2015spatial}
M.~Jaderberg, K.~Simonyan, A.~Zisserman, et~al.
\newblock Spatial transformer networks.
\newblock In {\em Advances in neural information processing systems}, pages
  2017--2025, 2015.

\bibitem{joo2015panoptic}
H.~Joo, H.~Liu, L.~Tan, L.~Gui, B.~Nabbe, I.~Matthews, T.~Kanade, S.~Nobuhara,
  and Y.~Sheikh.
\newblock Panoptic studio: A massively multiview system for social motion
  capture.
\newblock In {\em Proceedings of the IEEE International Conference on Computer
  Vision}, pages 3334--3342, 2015.

\bibitem{ke2018multi}
L.~Ke, M.-C. Chang, H.~Qi, and S.~Lyu.
\newblock Multi-scale structure-aware network for human pose estimation.
\newblock In {\em Proceedings of the European Conference on Computer Vision
  (ECCV)}, pages 713--728, 2018.

\bibitem{kong2019adaptive}
D.~Kong, Y.~Chen, H.~Ma, X.~Yan, and X.~Xie.
\newblock Adaptive graphical model network for 2d handpose estimation.
\newblock In {\em Proceedings of the British Machine Vision Conference
  ({BMVC})}, 2019.

\bibitem{lee2009multithreaded}
T.~Lee and T.~Hollerer.
\newblock Multithreaded hybrid feature tracking for markerless augmented
  reality.
\newblock {\em IEEE Transactions on Visualization and Computer Graphics},
  15(3):355--368, 2009.

\bibitem{lin2014microsoft}
T.-Y. Lin, M.~Maire, S.~Belongie, J.~Hays, P.~Perona, D.~Ramanan,
  P.~Doll{\'a}r, and C.~L. Zitnick.
\newblock Microsoft coco: Common objects in context.
\newblock In {\em European conference on computer vision}, pages 740--755.
  Springer, 2014.

\bibitem{mueller2018ganerated}
F.~Mueller, F.~Bernard, O.~Sotnychenko, D.~Mehta, S.~Sridhar, D.~Casas, and
  C.~Theobalt.
\newblock Ganerated hands for real-time 3d hand tracking from monocular rgb.
\newblock In {\em Proceedings of the IEEE Conference on Computer Vision and
  Pattern Recognition}, pages 49--59, 2018.

\bibitem{mueller2017real}
F.~Mueller, D.~Mehta, O.~Sotnychenko, S.~Sridhar, D.~Casas, and C.~Theobalt.
\newblock Real-time hand tracking under occlusion from an egocentric rgb-d
  sensor.
\newblock In {\em Proceedings of the IEEE International Conference on Computer
  Vision}, pages 1284--1293, 2017.

\bibitem{newell2016stacked}
A.~Newell, K.~Yang, and J.~Deng.
\newblock Stacked hourglass networks for human pose estimation.
\newblock In {\em European conference on computer vision}, pages 483--499.
  Springer, 2016.

\bibitem{panteleris2017back}
P.~Panteleris and A.~Argyros.
\newblock Back to rgb: 3d tracking of hands and hand-object interactions based
  on short-baseline stereo.
\newblock In {\em Proceedings of the IEEE International Conference on Computer
  Vision}, pages 575--584, 2017.

\bibitem{panteleris2018using}
P.~Panteleris, I.~Oikonomidis, and A.~Argyros.
\newblock Using a single rgb frame for real time 3d hand pose estimation in the
  wild.
\newblock In {\em 2018 IEEE Winter Conference on Applications of Computer
  Vision (WACV)}, pages 436--445. IEEE, 2018.

\bibitem{papandreou2017towards}
G.~Papandreou, T.~Zhu, N.~Kanazawa, A.~Toshev, J.~Tompson, C.~Bregler, and
  K.~Murphy.
\newblock Towards accurate multi-person pose estimation in the wild.
\newblock In {\em Proceedings of the IEEE Conference on Computer Vision and
  Pattern Recognition}, pages 4903--4911, 2017.

\bibitem{piumsomboon2013user}
T.~Piumsomboon, A.~Clark, M.~Billinghurst, and A.~Cockburn.
\newblock User-defined gestures for augmented reality.
\newblock In {\em IFIP Conference on Human-Computer Interaction}, pages
  282--299. Springer, 2013.

\bibitem{simon2017hand}
T.~Simon, H.~Joo, I.~Matthews, and Y.~Sheikh.
\newblock Hand keypoint detection in single images using multiview
  bootstrapping.
\newblock In {\em Proceedings of the IEEE conference on Computer Vision and
  Pattern Recognition}, pages 1145--1153, 2017.

\bibitem{song2017thin}
J.~Song, L.~Wang, L.~Van~Gool, and O.~Hilliges.
\newblock Thin-slicing network: A deep structured model for pose estimation in
  videos.
\newblock In {\em Proceedings of the IEEE Conference on Computer Vision and
  Pattern Recognition}, pages 4220--4229, 2017.

\bibitem{spurr2018cross}
A.~Spurr, J.~Song, S.~Park, and O.~Hilliges.
\newblock Cross-modal deep variational hand pose estimation.
\newblock In {\em Proceedings of the IEEE Conference on Computer Vision and
  Pattern Recognition}, pages 89--98, 2018.

\bibitem{sridhar2015investigating}
S.~Sridhar, A.~M. Feit, C.~Theobalt, and A.~Oulasvirta.
\newblock Investigating the dexterity of multi-finger input for mid-air text
  entry.
\newblock In {\em Proceedings of the 33rd Annual ACM Conference on Human
  Factors in Computing Systems}, pages 3643--3652. ACM, 2015.

\bibitem{sun2019deep}
K.~Sun, B.~Xiao, D.~Liu, and J.~Wang.
\newblock Deep high-resolution representation learning for human pose
  estimation.
\newblock {\em arXiv preprint arXiv:1902.09212}, 2019.

\bibitem{tompson2014joint}
J.~J. Tompson, A.~Jain, Y.~LeCun, and C.~Bregler.
\newblock Joint training of a convolutional network and a graphical model for
  human pose estimation.
\newblock In {\em Advances in neural information processing systems}, pages
  1799--1807, 2014.

\bibitem{toshev2014deeppose}
A.~Toshev and C.~Szegedy.
\newblock Deeppose: Human pose estimation via deep neural networks.
\newblock In {\em Proceedings of the IEEE conference on computer vision and
  pattern recognition}, pages 1653--1660, 2014.

\bibitem{wan2019self}
C.~Wan, T.~Probst, L.~V. Gool, and A.~Yao.
\newblock Self-supervised 3d hand pose estimation through training by fitting.
\newblock In {\em Proceedings of the IEEE Conference on Computer Vision and
  Pattern Recognition}, pages 10853--10862, 2019.

\bibitem{wan2018dense}
C.~Wan, T.~Probst, L.~Van~Gool, and A.~Yao.
\newblock Dense 3d regression for hand pose estimation.
\newblock In {\em Proceedings of the IEEE Conference on Computer Vision and
  Pattern Recognition}, pages 5147--5156, 2018.

\bibitem{wang2016deeply}
J.~Wang, Z.~Wei, T.~Zhang, and W.~Zeng.
\newblock Deeply-fused nets.
\newblock {\em arXiv preprint arXiv:1605.07716}, 2016.

\bibitem{wang2018mask}
Y.~Wang, C.~Peng, and Y.~Liu.
\newblock Mask-pose cascaded cnn for 2d hand pose estimation from single color
  image.
\newblock {\em IEEE Transactions on Circuits and Systems for Video Technology},
  2018.

\bibitem{gpa}
Z.~Wang, L.~Chen, S.~Rathore, D.~Shin, and C.~Fowlkes.
\newblock Geometric pose affordance: 3d human pose with scene constraints.
\newblock In {\em arxiv}, 2019.

\bibitem{wei2016convolutional}
S.-E. Wei, V.~Ramakrishna, T.~Kanade, and Y.~Sheikh.
\newblock Convolutional pose machines.
\newblock In {\em Proceedings of the IEEE Conference on Computer Vision and
  Pattern Recognition}, pages 4724--4732, 2016.

\bibitem{yang2016end}
W.~Yang, W.~Ouyang, H.~Li, and X.~Wang.
\newblock End-to-end learning of deformable mixture of parts and deep
  convolutional neural networks for human pose estimation.
\newblock In {\em Proceedings of the IEEE Conference on Computer Vision and
  Pattern Recognition}, pages 3073--3082, 2016.

\bibitem{yuan2018depth}
S.~Yuan, G.~Garcia-Hernando, B.~Stenger, G.~Moon, J.~Yong~Chang, K.~Mu~Lee,
  P.~Molchanov, J.~Kautz, S.~Honari, L.~Ge, et~al.
\newblock Depth-based 3d hand pose estimation: From current achievements to
  future goals.
\newblock In {\em Proceedings of the IEEE Conference on Computer Vision and
  Pattern Recognition}, pages 2636--2645, 2018.

\bibitem{zimmermann2017learning}
C.~Zimmermann and T.~Brox.
\newblock Learning to estimate 3d hand pose from single rgb images.
\newblock In {\em Proceedings of the IEEE International Conference on Computer
  Vision}, pages 4903--4911, 2017.

\end{thebibliography}
}

\end{document}